\documentclass[letterpaper, 10 pt, conference]{ieeeconf}  %

\IEEEoverridecommandlockouts                              %

\usepackage[T1]{fontenc} 
\usepackage[utf8]{inputenc}
\usepackage{amsmath}
\usepackage{amssymb} 
\usepackage{graphicx}
\usepackage[margin=1in]{geometry}
\usepackage{xcolor}
\usepackage[all]{nowidow}
\usepackage[hidelinks]{hyperref}
\usepackage{tikz}
\usepackage[ruled,vlined]{algorithm2e}
\DontPrintSemicolon
\usepackage{bm}
\usepackage{acronym}
\usepackage{svg}
\usepackage{flushend}
\usepackage{subcaption}
\usepackage[normalem]{ulem} %

\DeclareMathOperator*{\argmax}{arg\,max}

\DeclareMathOperator{\E}{\mathbb{E}}
\DeclareMathOperator*{\expval}{\mathbb{E}}

\newcommand{\set}[1]{\left\{#1\right\}}

\newcommand*{\btheta}{\bm{\theta}}
\newcommand*{\half}{\frac{1}{2}}

\title{\LARGE \bf
Sample Efficient Social Navigation\\ Using Inverse Reinforcement Learning
}

\author{Bobak H. Baghi$^{1}$ and Gregory Dudek$^{2}$%
\thanks{*This work supported by Samsung Electronics, and also benefited from the infrastructure provided by the National Science and Engineering Council of Canada (NSERC).}%
\thanks{$^{1}$Bobak H. Baghi and $^{2}$Gregory Dudek are with the School of Computer Science, McGill University. 3480 Rue University, Montreal, QC, Canada. H3A 2K6 
{\tt\small \{bobhb,dudek\}@cim.mcgill.ca}}
\thanks{$^{2}$Gregory Dudek is also with Samsung Electronics SAIC-Montreal. 1250 René-Lévesque, Montreal, QC, Canada. H3B 4K4 }
}

\usepackage{xcolor}

\begin{document}

\acrodef{SAC}{soft actor critic}
\acrodef{RL}{reinforcement learning}
\acrodef{IRL}{inverse reinforcement learning}
\acrodef{MEDIRL}{maximum entropy deep inverse reinforcement learning}
\acrodef{MaxEnt}{maximum entropy}
\acrodef{GCL}{guided cost learning}
\acrodef{MLE}{maximum likelihood estimation}
\acrodef{det-MEDIRL}{deterministic MEDIRL}
\acrodef{RMSE}{root mean square error}
\acrodef{ROI}{region of interest}
\acrodef{LfO}{learning from observation}
\acrodef{LfD}{learning from demonstration}
\acrodef{RCAP}{reciprocal collision avoidance for pedestrians}
\acrodef{ORCA}{optimal reciprocal velocity obstacles}
\acrodef{RRT}{rapidly exploring random trees}
\acrodef{SVF}{state visitation frequency}
\acrodef{BIRL}{Bayesian IRL}

\maketitle
\thispagestyle{empty}
\pagestyle{empty}

\begin{abstract}

In this paper, we present an algorithm to efficiently learn socially-compliant navigation policies from observations of human trajectories.
As mobile robots come to inhabit and traffic social spaces, they must account for social cues and behave in a socially compliant manner.
We focus on learning such cues from examples.
We describe an inverse reinforcement learning based algorithm which learns from human trajectory observations without knowing their specific actions. We increase the sample-efficiency of our approach over alternative methods by leveraging the notion of 
a replay buffer (found in many off-policy reinforcement learning methods) to eliminate the additional sample complexity associated with inverse reinforcement learning.
We evaluate our method by training agents using publicly available pedestrian motion data sets and compare it to related methods. We show that our approach yields better performance while also decreasing training time and sample complexity.

\end{abstract}

\section{INTRODUCTION}
In this paper, we present a new algorithmic formulation to allow 
a robot to learn a navigation policy directly by
observing human examples.   We use inverse reinforcement learning, but modify the formulation to provide advantages over prior work (including our own).

Recent work has demonstrated the potential of mobile robots that operate in human-inhabited social spaces, such as 
service robots for the elderly %
\cite{jayawardenaDeploymentServiceRobot2010}, as shopping assistants \cite{grossTOOMASInteractiveShopping2009}, 
and in the office environment \cite{mitsunagaRobovieIVCommunicationRobot2006}.

Robotic navigation is a well-understood area of robotics and contains rich literature concerning efficient robotic movement in challenging environments under various constraints. For the subproblem referred to as ``social navigation'', however, 
the need to account for socially appropriate behavior is often as important as efficiency~\cite{guyModelingCollisionAvoidance2010}.
This reflects the fact that  integrating mobile robots into human social spaces goes beyond simple obstacle avoidance and actively acts in a way that makes humans 
nearby comfortable.
Such socially aware policies act according to the socially acceptable navigation rules and closely mimic the behavior exhibited by humans.

Unfortunately, the guidelines for suitable social navigation are not readily codified algorithmically (despite several attempts), often leading to fragile
threshold-dependent behavior, and are quite context-dependent.  As such, it behooves a system to become able to infer appropriate socially-aware policies from observational data.  One approach to this is to mimic the behavior of humans in the same context.

While \ac{RL} can lead to powerful solutions when the reward structure for the problem is well understood, social navigation depends on a range of poorly understood cues that may depend on the scenario. Indeed, the problem of reward engineering is a pivotal challenge to obtaining desired behavior out of any \ac{RL} based controller. This limits the modeling power of any hand-crafted reward function for the sake of socially compliant navigation.

Ideally, the reward structure of socially compliant behavior would be learned from demonstrations of human pedestrians who already exhibit such socially compliant behavior. As such, \ac{IRL} is attractive since it allows us to infer suitable rewards from raw features based on observations. A policy trained on these learned rewards using \ac{RL} yields a suitable socially compliant navigation controller.

The \ac{IRL} process, however, is typically computationally expensive and sample inefficient, partly due to the amount and quality of diverse demonstration data needed and due to the ill-posed nature of its mathematical formulation. These obstacles towards practical inverse reinforcement learning for social navigation are compounded by the difficulty of defining useful performance metrics to judge the performance of resulting algorithms and models (unsurprisingly, since robust and clear performance metrics could be used as optimization objectives).  

In this paper, we present a model-free and sample efficient deep inverse reinforcement learning based navigation pipeline that can produce artificial agents that learn social navigation from observation. 
By leveraging existing trajectories within the replay buffer of off-policy \ac{RL} algorithms, we eliminate the need for further environment interaction during the reward learning phase, yielding better sample efficiency and faster training.
We evaluate the performance of our approach on metrics capturing desirable aspects of social navigation as well as its sample efficiency and compare against other approaches in the literature.

\section{RELATED WORK}

A natural blueprint for human-aware navigation is to extend classical obstacle avoidance and navigation planning algorithms to fit the constraints of this new domain. 
One such approach is the reciprocal collision avoidance for pedestrians (RCAP) algorithm \cite{guyModelingCollisionAvoidance2010} which extends ORCA \cite{vandenbergReciprocalNBodyCollision2011a} by taking various human factors such as reaction time, physical constraints, and personal space into account.
Müller et al.~\cite{mullerSociallyInspiredMotion2008} extend A* \cite{hartFormalBasisHeuristic1968a} by identifying and following a leader until the social obstacle is overcome.
Svenstrup et al.~\cite{svenstrupTrajectoryPlanningRobots2010} propose a social path planning algorithm that models humans as dynamic obstacles with inviolable social zones enforced by a dynamic potential field optimized by \ac{RRT} \cite{lavalleRapidlyexploringRandomTrees1998}.

Pioneering work from Helbing and Molnar~\cite{helbingSocialForceModel1998} models human behavior using repulsive forces exerted between agents in an environment and an attractive force towards the goal position. Since the model parameters require careful tuning, some researchers have used camera footage for calibration~\cite{helbingPedestrianCrowdEvacuation2010,johanssonSpecificationSocialForce2007}, while others have made use of genetic algorithms to the same end~\cite{ferrerRobotCompanionSocialforce2013}. These approaches, while succinct, cannot capture context-dependencies and the subtle behaviors humans exhibit.

Chen et al.~\cite{chenDecentralizedNoncommunicatingMultiagent2017} make use of \ac{RL} to train a navigational policy 
using a hand-crafted reward function that incentivizes reaching the destination and penalizes undesirable collisions. This was improved upon by Everett et al.~\cite{everettMotionPlanningDynamic2018c} which leveraged GPU processing to train multiple simulated agents in parallel, and an LSTM network \cite{hochreiterLongShortTermMemory1997} to compress the state information of an arbitrary number of adjacent agents into a fixed-size vector. \cite{chenSociallyAwareMotion2017a} demonstrate that the \ac{RL} approach has the ability to learn socially compliant behavior (specifically, a social norm of passing by the right side) using a carefully designed reward function that depends on
the situational dynamics. 

In practice, 
the choice of reward function can dramatically impact learning rate and quality. \ac{IRL} has thus been used to learn reward functions directly from observations of human pedestrian traffic.
A widely-used \ac{IRL} framework is \ac{MaxEnt} \ac{IRL} \cite{ziebartMaximumEntropyInverse2008} which allows the parameters of a reward function to be optimized using \ac{MLE}.
Kitani et al.~\cite{kitaniActivityForecasting2012} apply a modified \ac{MaxEnt} \ac{IRL} \cite{ziebartMaximumEntropyInverse2008} model to forecast pedestrian trajectories from noisy camera data which can also double as a navigation controller.

Vasquez et al.~\cite{vasquezInverseReinforcementLearning2014} investigate composite features leveraging the motion information of nearby pedestrians and demonstrate that \ac{IRL} learned reward parameters are superior to manually tuned ones.
Henry et al.~\cite{henryLearningNavigateCrowded2010} use Gaussian processes (GP) to estimate environment features from incomplete data and use rewards learned from \ac{MaxEnt} \ac{IRL} as A* heuristics for path planning, quantitatively demonstrating a more human-like quality to the chosen trajectories.

Kretzschmar et al.~\cite{kretzschmarSociallyCompliantMobile2016}, building on previous work~\cite{kudererFeatureBasedPredictionTrajectories,kretzschmarLearningPredictTrajectories2014}, apply a maximum entropy approach to learn a joint trajectory distribution of all navigating agents in an environment, including the robot itself. 
Inference over the spline-based trajectory representation is made possible by heuristically sampling high probability regions from certain defined homotopy classes.
This approach is tested on a real robot in an overtake scenario~\cite{kudererFeatureBasedPredictionTrajectories} and demonstration data collected from four pedestrians in an enclosed environment~\cite{kretzschmarLearningPredictTrajectories2014} and a dataset~\cite{pellegriniYouLlNever2009} of three to five pedestrians. While learning a joint distribution over all agents allows for high-quality inference, it does not scale to moderately populated settings (e.g. a few dozen agents).

An alternative probabilistic approach to \ac{IRL}, \ac{BIRL}, is proposed by Ramachandran et al.~\cite{ramachandranBayesianInverseReinforcement2007} where the reward is modeled as a random variable vector that dictates the distribution of expert states and actions. The distribution over the rewards which best explains expert trajectories is then inferred.

Okal et al.~\cite{okalLearningSociallyNormative2016} use a modified \ac{BIRL} approach using control graphs \cite{neumannEfficientContinuousTimeReinforcement2007a} to tractably capture continuous state information via macrostates in a graph-based representation. Since a graph-based representation is learned, local controllers can be used in conjunction with global planning algorithms such as \ac{RRT} or A* using the learned rewards for navigation, as the authors demonstrate both in a simulation of artificial pedestrians and via a real robot. Kim et al.~\cite{kimSociallyAdaptivePath2016} use \ac{BIRL} to learn a linear reward function over features extracted from RGB-D cameras on a robotic wheelchair, which are then used for path planning purposes in crowded environments.

The linear reward function used by \ac{MaxEnt} \ac{IRL} \cite{ziebartMaximumEntropyInverse2008} struggles with complex non-linear rewards. To address this, Wulfmeier et. al~\cite{wulfmeierMaximumEntropyDeep2016} introduce \ac{MEDIRL}, extending \ac{MaxEnt} \ac{IRL} to deep neural network based reward functions.
Fahad et al.~\cite{fahadLearningHowPedestrians2018} leverage \ac{MEDIRL} and velocity-augmented SAM features \cite{alahiSociallyAwareLargeScaleCrowd2014} to train a navigation policy on pedestrian data in a grid world setting. Konar et al.~\cite{konarLearningGoalConditioned2021} propose \ac{det-MEDIRL}, which efficiently learns goal conditioned navigation policies using a risk-based feature representation that segments the agent's surrounding pedestrians based on the posed collision risk.
These approaches require many costly policy roll-outs at each iteration of the reward function optimization, which greatly slows down training.

\section{INVERSE REINFORCEMENT LEARNING}

We begin by outlining IRL-based learning for social navigation, and then describe how we enhance it.
A Markov decision process (MDP) $\mathcal{M}$ is defined by the tuple $( \mathcal{S}, \mathcal{A}, \mathcal{T}, R,\gamma, p_0)$ where $\mathcal{S}$ is the set of possible states, $\mathcal{A}$ is the set of possible actions, transition probabilities $\mathcal{T}(s,a,s')=P(s'=s_{t+1}|s = s_t, a = a_t)$
modelling the probability of transitioning from state $s_t$ to $s_{t+1}$ by taking action $a_t$.
$R: \mathcal{S} \mapsto \mathbb{R}$ is the reward function, $\gamma$ is the discounting factor, and $p_0=P(s=s_0)$ is the probability distribution of possible initial states. 

An observed trajectory $\tau$ of length $t$ is any valid ordered sequence of states where $s_i \in \mathcal{S}$.
\begin{gather*}
    \tau_{obs} = \set{s_0, s_1, \dots, s_t}
\end{gather*}
In this work, we are \ac{LfO} as opposed to \ac{LfD} as we do not assume access to the actions of the expert. This distinction is important since, among other factors, action selection includes non-determinism and often a significantly different action space between expert and robot. Due to the lack of action information, \ac{LfO} is a harder problem to solve~\cite{yangImitationLearningObservations2019a}.

A (stochastic) policy is a mapping $\pi: \mathcal{S} \times \mathcal{A} \mapsto [0,1]$. 
For an agent which obtains rewards through environment interaction,
reinforcement learning (RL) is the process of finding an optimal policy $\pi^*$ from the set of admissible policies $\Pi$ that maximizes the discounted total returns.
\begin{equation}
    \pi^* = \argmax_{\pi\in\Pi} \E_{\tau\sim\pi} \left[\sum_{s_t,a_t \in \tau} \gamma^t R(s_t,a_t,s'_t)\right]
\end{equation}

The goal of inverse reinforcement learning (IRL) is to determine the optimal reward function $R^*$ which an expert policy $\pi_E$ is
presumably optimizing. That can be seen as a reward-based explanation for the observed behavior.  In our case, this would be computed over trajectories $D = \{ \tau_1, \tau_2, \dots, \tau_M \}$ from the expert policy.

We use the \ac{MaxEnt} \ac{IRL} framework \cite{ziebartMaximumEntropyInverse2008, ziebartModelingInteractionPrinciple2010}, which assumes a maximum entropy distribution over the trajectory space
\begin{gather}
    P(\tau| \btheta) = \frac{1}{Z(\btheta)}\exp(R_{\btheta}(\tau))
\end{gather}
where $R_{\btheta}$ is a reward function with parameters $\btheta$, and $Z(\btheta)$ is the partition function. With a differentiable reward function and a convergent partition function, \ac{MLE} can be used to fit this distribution to a set of expert observation trajectories $D_E$. 
While prior work derives a \ac{SVF} based derivative %
\cite{konarLearningGoalConditioned2021,fahadLearningHowPedestrians2018},
we use importance sampling to estimate the partition function following \ac{GCL} \cite{finnGuidedCostLearning2016}
\begin{align}
     Z(\btheta) &= \int_\tau \exp(R_\theta(\tau))d\tau \approx \expval_{\tau \sim q(\tau)} \frac{\exp(R_\theta(\tau))}{q(\tau)}
\end{align}
With this approximation, the \ac{MaxEnt} log-likelihood becomes
\begin{gather}
    L(\btheta) \approx \expval_{\tau_E \in D_E} R_{\btheta}(\tau_E) - \log \expval_{\tau \sim D_q} \frac{\exp(R_{\btheta}(\tau))}{q(\tau)}
    \label{eq:GCL-likelihood}
\end{gather}

\ac{GCL} uses a policy trained on the current rewards as the sampling policy $q(\tau)$, thus minimizing the variance of the importance sampling estimator \cite{finnGuidedCostLearning2016}
The training of the policy is interleaved with the rewards function, removing \ac{RL} as inner-loop component and greatly accelerating training. In practice, however, simply using the latest iteration policy is impractical due to the initially poor coverage. As such, a mixture distribution 
$\mu(\tau) = \half\pi(\tau) + \half\hat{q}(\tau)$ is used as sampling distribution, where $\hat{q}(\tau)$ is a crude approximate distribution over the expert trajectories \cite{finnGuidedCostLearning2016, finnConnectionGenerativeAdversarial2016}.

The \ac{MaxEnt} \ac{IRL} framework learns rewards that induce policies that match the expert features. For social navigation, we use risk features~\cite{konarLearningGoalConditioned2021} that encode relative pedestrian locations and change in velocity which should lead to a less intrusive policy with smoother trajectories when applied to social navigation, as well as features orienting the agent towards the goal position, which should lead to collision-free navigation.

\section{Improving Sample Efficiency}
An important technique in off-policy \ac{RL} is the use of experience replay buffers~\cite{linSelfimprovingReactiveAgents1992} to boost sample efficiency by way of re-using old experiences for training. In our method, which we call ``ReplayIRL'', we make use of \ac{SAC}, a recent sample-efficient \ac{RL} algorithm~\cite{haarnojaSoftActorCriticOffPolicy2018, haarnojaSoftActorCriticAlgorithms2018}. We further increase the sample efficiency of our algorithm by reusing the trajectories collected in the replay buffer as samples for the importance sampling calculation of the reward log-likelihood, an approach that was previously employed in the related imitation learning domain \cite{kostrikov2018discriminatoractorcritic}.

We use an alternative formulation of the \ac{GCL} log-likelihood \eqref{eq:GCL-likelihood} by Fu et al.~\cite{fuLearningRobustRewards2017} as it is more stable, and includes discounting while using a state-only reward function formulation.
\begin{gather}
    L_{obs}(\btheta) = \expval_{\tau\sim D_E} \left[ \sum_{s_t\in\tau}  \gamma^t R_{\btheta}(s_t) \right]  \label{eq:airl-expert} \\
    L_{IS}(\btheta) = \expval_{\tau\sim \mu} \left[  \sum_{s_t, a_t\in\tau} \log (e^{\gamma^t R_{\btheta}(s_t)} + \pi(a_t|s_t)) \right] \label{eq:airl-IS}\\
    L(\btheta) = L_{obs}(\btheta) - L_{IS}(\btheta)  \label{eq:airl-full-obj}
\end{gather}

We modify the above equations to sample from a replay buffer $B$ instead of performing policy roll-outs. Additionally, we share the expert trajectory samples for the calculation of both $L_{obs}$ and $L_{IS}$. This results in a mixture distribution 
\begin{gather}
    \hat\mu(\tau) = \half D_E(\tau) + \half B(\tau)
\end{gather}
where we slightly overload notation by treating the replay buffer $B$ and expert trajectories $D_E$ as distributions. In practice, we simply uniformly sample trajectories from both collections.

Note that in \eqref{eq:airl-IS}, the quantity $\pi(a|s)$ cannot be evaluated for expert trajectories as their actions are unknown. 
We found that letting $\pi(a|s) = \frac{1}{\sigma_T\sqrt{2\pi}}$, a constant value, is sufficient for training. This crude approximation corresponds to the density of a Gaussian policy with target standard deviation $\sigma_T$ at the mean value.

Following the above modifications, the importance sampling portion of the algorithm becomes
\begin{gather}
    A(\btheta, \tau) =
    \begin{cases}
    \sum\limits_{s_t, a_t\in\tau} \log (e^{\gamma^t R_{\btheta}(s_t)}+\pi(a_t|s_t)) 
    & \tau \in B \\
    \sum\limits_{s_t, a_t\in\tau} \log (e^{\gamma^t R_{\btheta}(s_t)} + \frac{1}{\sigma_T\sqrt{2\pi}})
    & \tau \in D_E
    \end{cases} \\
    \hat L_{IS}(\btheta, \tau_B) = \expval_{\tau \in \tau_B} \left[ A(\btheta, \tau) \right] 
    \label{eq:real-IS} \\ 
    \hat L_{obs}(\btheta, \tau_E) = \expval_{\tau\in \tau_E} \left[ \sum_{s_t\in\tau}  \gamma^t R_{\btheta}(s_t)
    \right]
    \label{eq:real-obs}
\end{gather}
where we use $\tau_B$ and $\tau_E$ as placeholders for appropriately sampled sets of trajectories passed to these functions, since in practice these trajectories are sampled before computing the above quantities. Finally, we use \ac{SAC} with automatic entropy tuning \cite{haarnojaSoftActorCriticAlgorithms2018}, and update the rewards of the sampled batch, ensuring that \ac{SAC} will perform updates with the up-to-date reward function. The full algorithm can be found in \autoref{alg:replayirl}.

\begin{algorithm}[tbhp]
\caption{Replay IRL}
\label{alg:replayirl}

\SetKwInOut{Input}{Input}
\SetKwInOut{Output}{Output}

\Input{expert demonstrations $D_E$ \newline
maximum iterations $M$\newline
training intervals  $i_{RL}$, $i_{IRL}$ %
}
\Output{optimized parameter $\btheta$ and policy $\pi$}

\SetKwComment{Comment}{$\triangleright$\ }{}
\SetKwFunction{Backprop}{Backprop}
\SetKwFunction{UpdateWeight}{UpdateWeight}
\SetKwFunction{SolveMDP}{SolveMDP}
\SetKwFunction{UpdateReward}{UpdateReward}
\SetKwFunction{UpdatePolicy}{UpdatePolicy}
\SetKwFunction{ReplayIRL}{ReplayIRL}
\BlankLine

\SetKwProg{irlalg}{Procedure}{}{}
\irlalg{\UpdateReward{}}{
    sample $n_E$ expert trajectories $\tau_E \sim D_E$ \;
    sample $n_B$ replay buffer trajectories $\tau_B \sim B$ \;
    $\tau_B \gets \tau_B \cup \tau_E$ \;
    $\hat L(\btheta) \gets \nabla_{\btheta} \left(\hat L_{obs}(\btheta, \tau_E) - \hat L_{IS}(\btheta, \tau_B) \right)$ \;
    $\btheta \gets \btheta - \alpha \nabla_{\btheta} L(\btheta)$ \;
}

\SetKwProg{rlalg}{Procedure}{}{}
\rlalg{\UpdatePolicy{}}{
    execute $\pi(a|s)$ to obtain $(s_t,a_t,s_{t+1})$ \;
    $B \gets B \cup (s_t,a_t,s_{t+1})$ \;
    $B_R \gets$ sample batch $(s_t,a_t,s_{t+1},R_{\btheta}(s_{t+1}))$ \;
    update $\pi$ using SAC iteration using $B_R$ \;
}

\BlankLine
\Begin{
 Initialize $\bm{\theta} \gets \bm{\theta}_0$, $B \gets\emptyset$ \; \DontPrintSemicolon 
 \For{$m=1$ \KwTo $M$}{
    \If{$m \mod i_{RL}$}{
        \UpdatePolicy{}
    }
    \If{$m \mod i_{IRL}$}{
        \UpdateReward{}
    }
 }
}
\end{algorithm}

\section{Experiments}
We compare ReplayIRL
to a recent deep \ac{IRL} social navigation method,  \ac{det-MEDIRL}~\cite{konarLearningGoalConditioned2021}, as well as \ac{SAC} trained on hand-crafted rewards. 
We use the UCY dataset \cite{lernerCrowdsExample2007} both as expert trajectories for training as well as for pedestrian placement in our simulated environment. 

\subsection{Pedestrian Dataset Processing}
The UCY dataset \cite{lernerCrowdsExample2007} is composed of five subsets students001, students003, zara01, zara02, and zara03. We concatenate the zara01-03 subsets into one, resulting in three distinct subsets students001, students003, and zara. The former two are from a mid-density public space in a university campus, while the latter is a lower-density sidewalk. The per-frame pedestrian positions and approximate gaze direction are provided, of which we only make use of the former.

We make use of an approximate homography to transform the overhead view from which the original data was collected into a top-down view. As a side effect, this also allows us to roughly convert the pixel space coordinates to world space.

We also removed inaccessible regions, trajectories that exhibited collisions, and
other anomalous observations.
The final pedestrian count is 365, 341, and 443 pedestrians in the students001, students003, and zara data sets respectively.

\begin{figure}
\centering\centering
\hfill
\begin{subfigure}[tbhp]{0.49\columnwidth}
\centering
    \includegraphics[width=0.93\columnwidth]{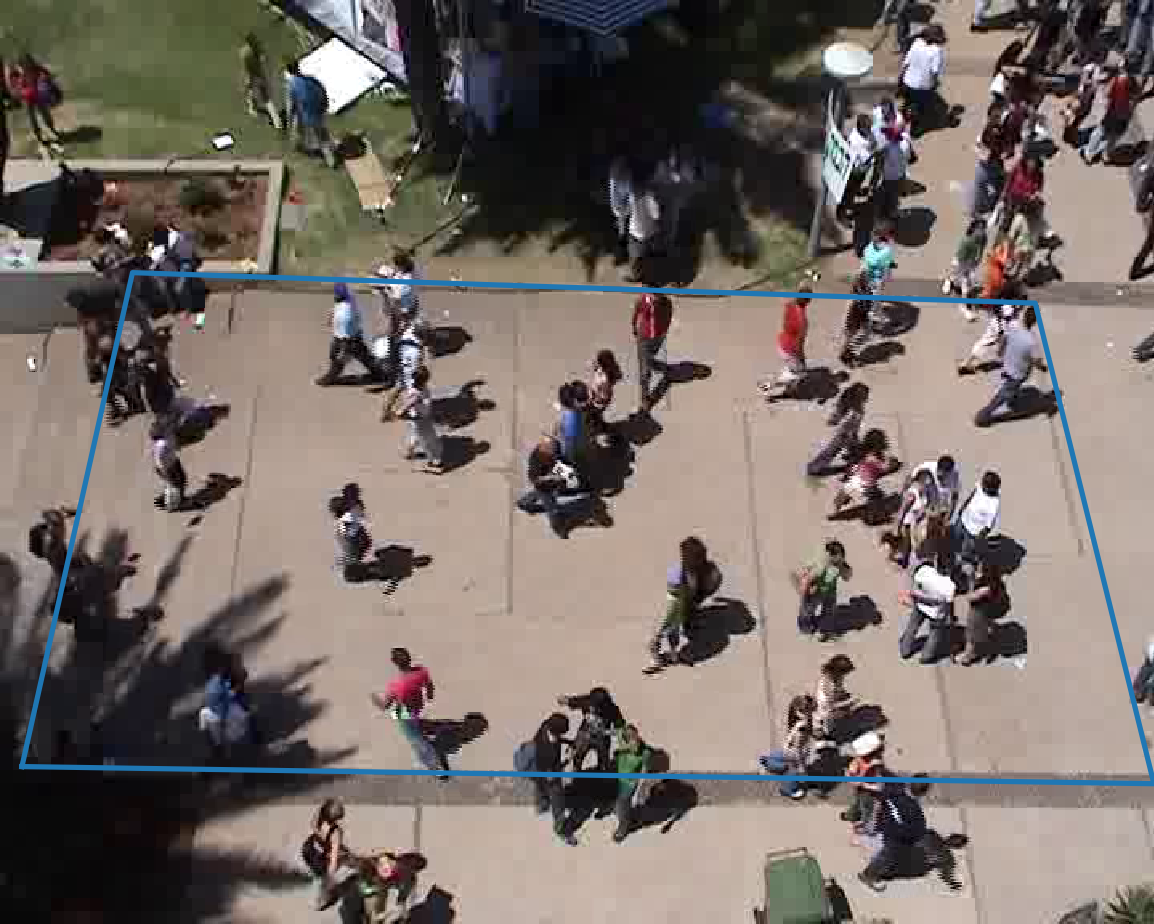}
    \caption{students001 subset.}
    \label{fig:students-unwarped-constraints}
\end{subfigure}
\hfill
\begin{subfigure}[tbhp]{0.49\columnwidth}
\centering
    \includegraphics[width=0.8\columnwidth]{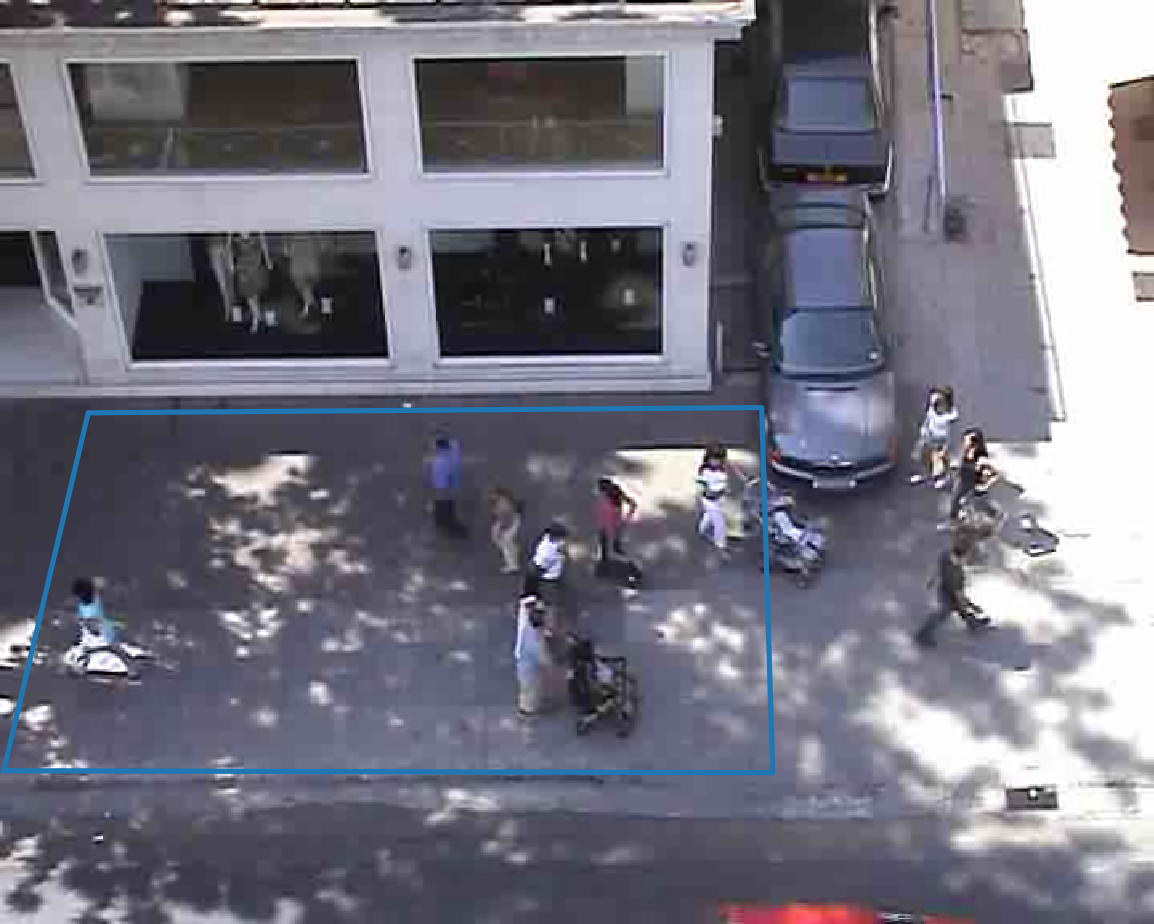}
    \caption{Zara subset.}
    \label{fig:zara-unwarped-constraints}
\end{subfigure}

\begin{subfigure}[tbhp]{0.49\columnwidth}
    \centering
    \includegraphics[height=6em]{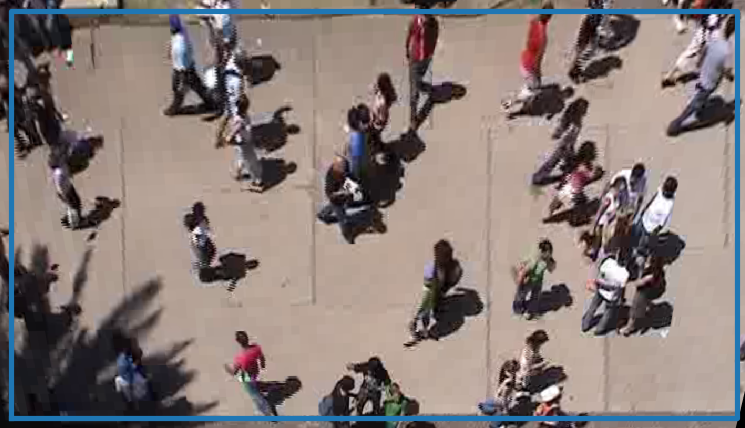}
    \caption{students001 subset.}
    \label{fig:students-warped-constraints}
\end{subfigure}
\hfill
\begin{subfigure}[tbhp]{0.49\columnwidth}
    \centering
    \includegraphics[height=6em]{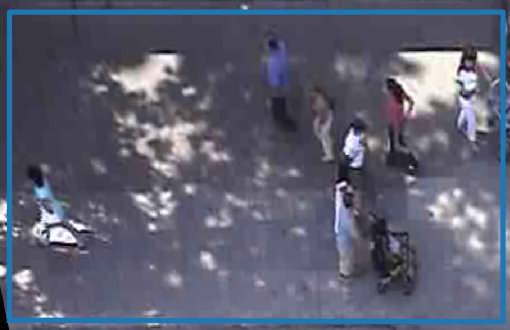}
    \caption{Zara subset.}(x
    \label{fig:zara-warped-constraints}
\end{subfigure}

\caption{(a) and (b) show \ac{ROI} with blue bounding box in UCY locations {\cite{lernerCrowdsExample2007}}. (c) and (d) show corresponding corrected top-down approximation via homography.}
\end{figure}

\subsection{Simulation Environment}
We implement a simulation environment to allow our agent to train in a crowded environment. 
We model pedestrians as discs with a diameter of $0.2\:m$ and position them 
according to the dataset.
The continuous action space consists of the target forward speed in the range $[0, 1.5] \:m/s$, and change in heading angle limited to the range $[0, \frac{3\pi}{2}]\:\text{deg}/s$.
The agent is deterministically displaced based on its resulting velocity vector. Each simulator time step advances time by $0.04 s$. For states, we compute risk features \cite{konarLearningGoalConditioned2021} which encode collision risk from nearby pedestrians. We use $0.65\;m$ and $1.0\;m$ as the radius of the inner and outer circles.

When each episode starts, a pedestrian is removed from the simulation for the duration of the experiment and replaced by the agent. The goal position of the agent is the final position of the removed pedestrian. The episode terminates when the agent is in the $10\;cm$ radius of the goal coordinates or after a $40\:s$ (1000 time step) timeout.

\subsection{Training details} \label{subsec:training-details}
We use a 2-layer fully connected deep neural network with 256 and 128 neurons per layer to represent the policy and reward networks respectively. We make use of the ReLU activation function \cite{nairRectifiedLinearUnits2010} in both networks, and make use of the AdamW optimizer \cite{loshchilovDecoupledWeightDecay2019} with $l2$ weight decay of $10^{-4}$, learning rate of $10^{-4}$ for the reward network and $3\cdot10^{-4}$ for SAC, learning rate decay $\gamma=0.9999$, reward update interval $i_{IRL}=3$, policy update interval $i_{RL}=1$, number of expert and policy samples $n_E=n_B=16$, 512 replay buffer samples for SAC. Other hyperparameters for \ac{SAC} are the same as in \cite{haarnojaSoftActorCriticAlgorithms2018}.

We compare our algorithm with another deep \ac{IRL} based social navigation algorithm, \ac{det-MEDIRL}. %
We slightly modify \ac{det-MEDIRL} to use \ac{SAC} and batch gradient descent with the same batch size as ReplayIRL. 
This reduces the training time of \ac{det-MEDIRL} from 72 to roughly 28 wall-time hours and while increasing sample efficiency without affecting the original performance (see \autoref{sec:results}). 

We also compare against a baseline \ac{SAC} agent with hand-crafted rewards based on \cite{everettMotionPlanningDynamic2018c} with a shaped reward to help the agent reach the goal. The reward structure can be decomposed into the following components: 
\begin{gather}
    R_{\text{approach}} = 0.1 \times \frac{\Delta x \cdot d_{rg} }{||  d_{rg} ||_2 }  \\
        R_G = 
    \begin{cases}
    1 & \text{if} ~ ||x_r - x_g||_2 \leq r_g \\
    0 & \text{else}
    \end{cases} \\
    R_{\text{col}} = 
    \begin{cases}
    -1 & \text{if} ~ d_{min} < 2 \cdot r_{ped} \\
    -0.01 \times d_{min} & \text{if} ~ d_{min} < 4 \cdot r_{ped}
    \end{cases}
\end{gather}
Where $x_t$ and $x_{t-1}$ are the current and previous robot locations respectively, $\Delta x = x_t - x_{t-1}$ is the displacement vector, $x_g$ and $r_g$ are the goal location and radius respectively, $d_{min}$ is the distance to the nearest pedestrian, and $r_{ped}$ is the radius of the disc that models pedestrians.
The total reward is then calculated $R = R_\text{approach} + R_\text{col} + R_G$.

We train ReplayIRL for $1.3\cdot 10^5$ iterations, \ac{det-MEDIRL} for 2,000 iterations, and SAC for $4\cdot 10^5$ iterations. Every training experiment is carried out over 5 random seeds. All methods are trained on the student001 dataset and evaluated on student003 and zara datasets.

\section{RESULTS} \label{sec:results}
We evaluate the trained social navigation policies based on metrics that would, together, capture desirable aspects of social navigation.
For every pedestrian in the test datasets (student003 and zara), we collect a trajectory by replacing said pedestrian with the agent. We repeat this for every starting position (i.e. every pedestrian). 
Furthermore, we collect metrics comparing the sample efficiency of the \ac{IRL} based algorithms. 
Error bars and shaded areas represent 95\% confidence interval in all figures.

\subsection{Performance Results}

\subsubsection{Proxemic Intrusions}
Informed by proxemics \cite{hallHandbookProxemicResearch1974}, we record the number of intimate and personal space intrusions. At each evaluation time step, for every pedestrian, the corresponding intrusion count is incremented should their mutual distance fall within the thresholds $[0,0.5]\;m$ and $(0.5,1.2]\;m$ for intimate and personal distance respectively.

\begin{figure} 
\centering\centering
\begin{subfigure}[tbhp]{0.49\columnwidth}
    \includegraphics[width=\columnwidth]{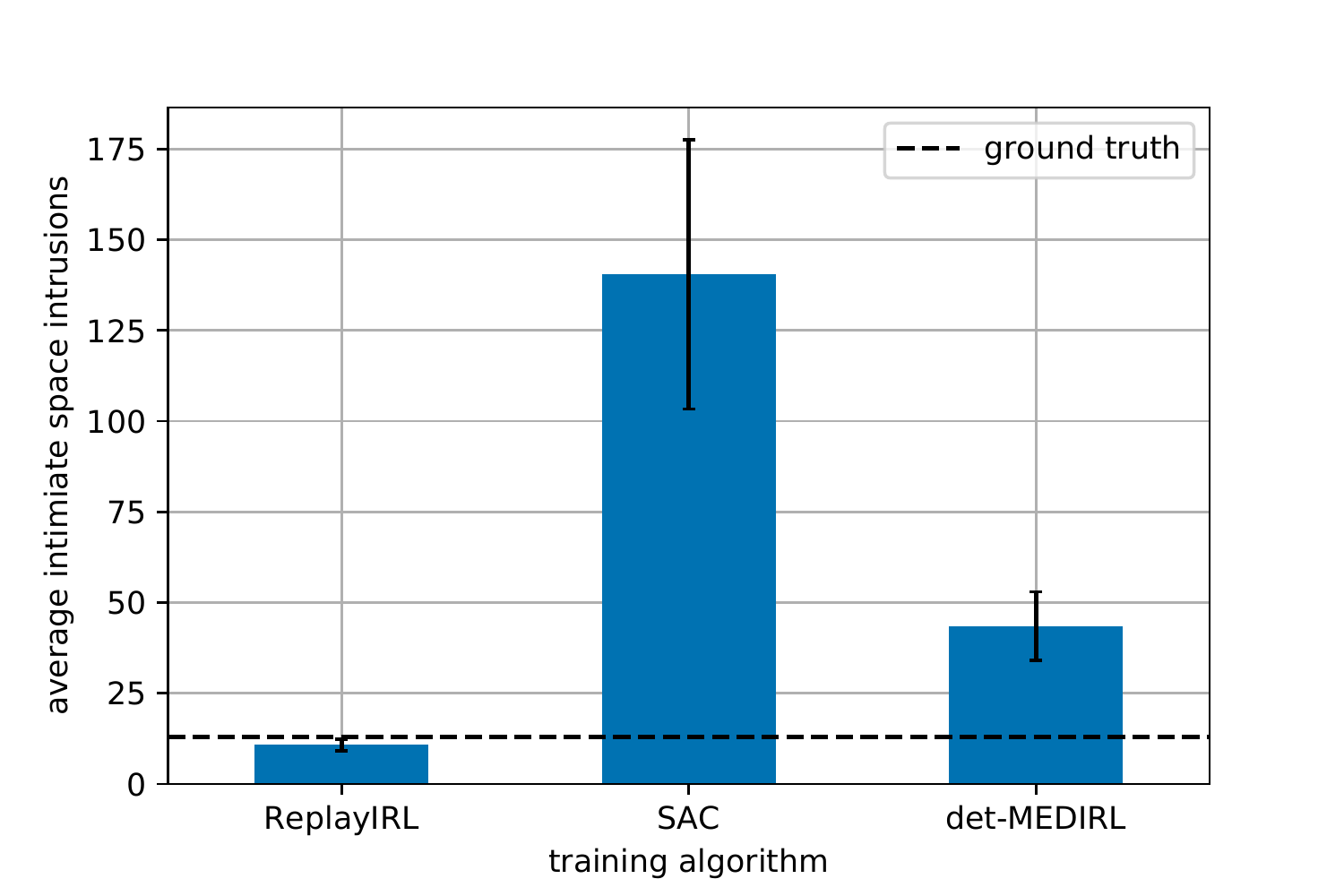}
    \caption{students003 subset.}
\end{subfigure}
\hfill
\begin{subfigure}[tbhp]{0.49\columnwidth}
    \includegraphics[width=\columnwidth]{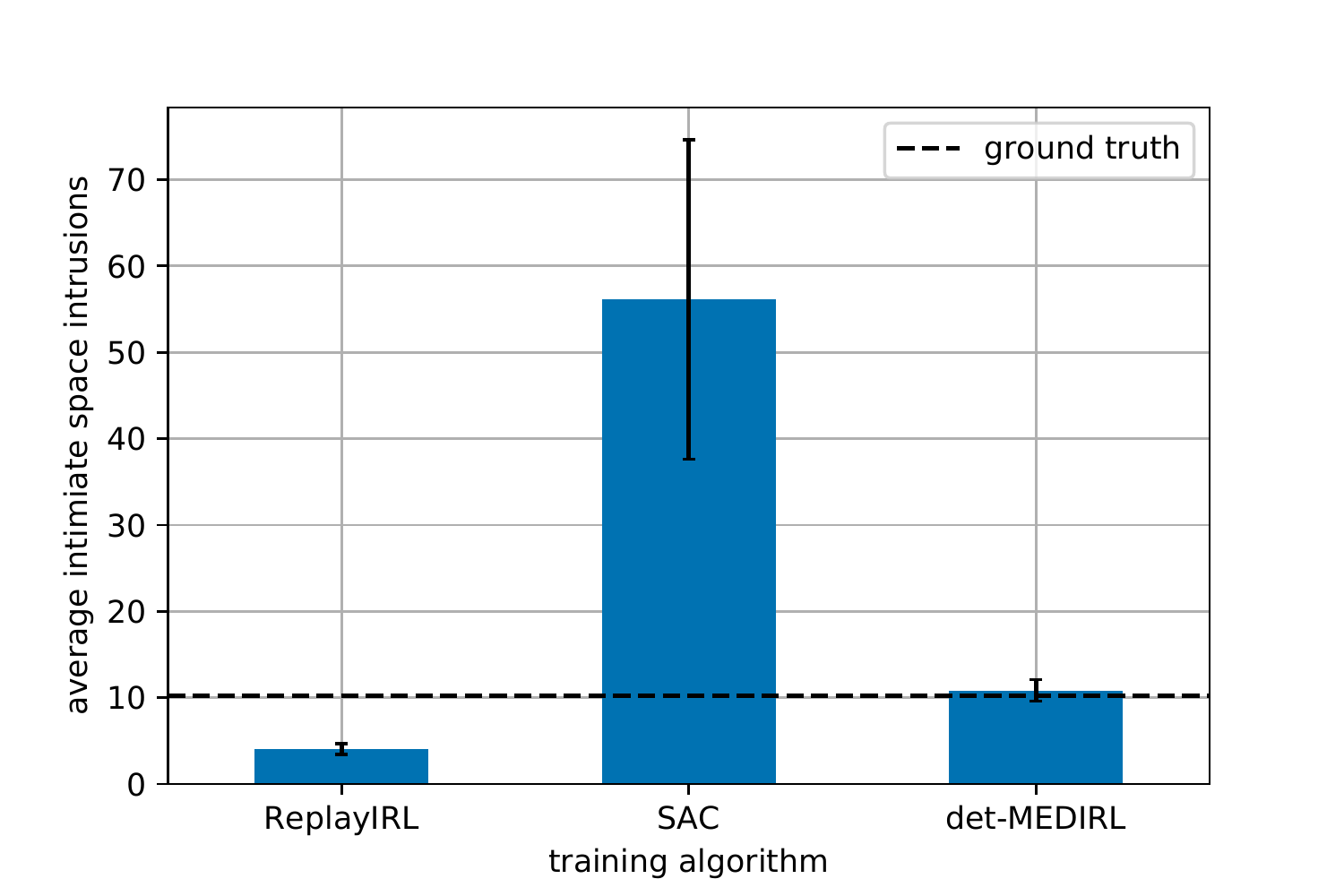}
    \caption{Zara subset.}
\end{subfigure}
\begin{subfigure}[tbhp]{0.49\columnwidth}
    \includegraphics[width=\columnwidth]{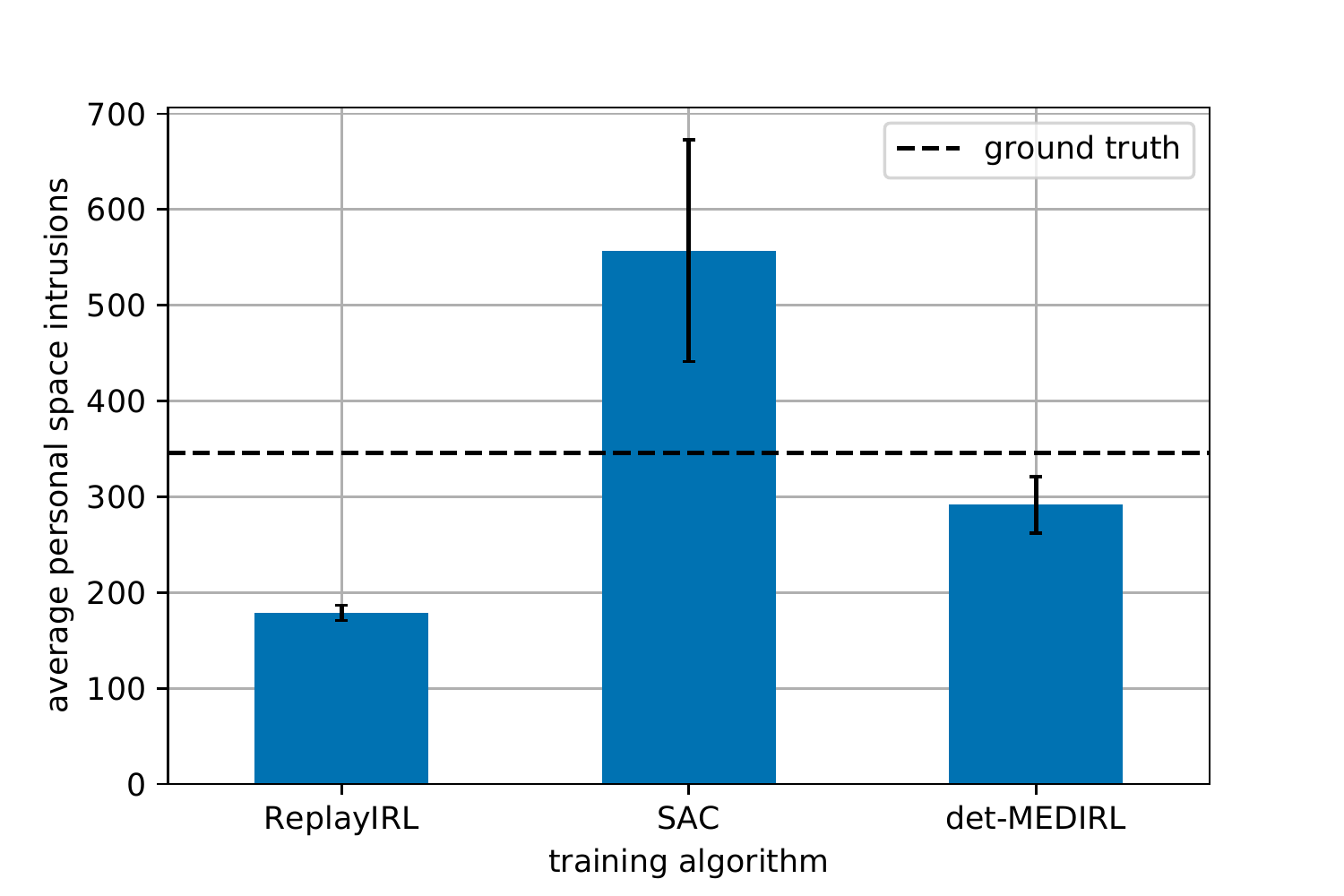}
    \caption{students003 subset.}
\end{subfigure}
\hfill
\begin{subfigure}[tbhp]{0.49\columnwidth}
    \includegraphics[width=\columnwidth]{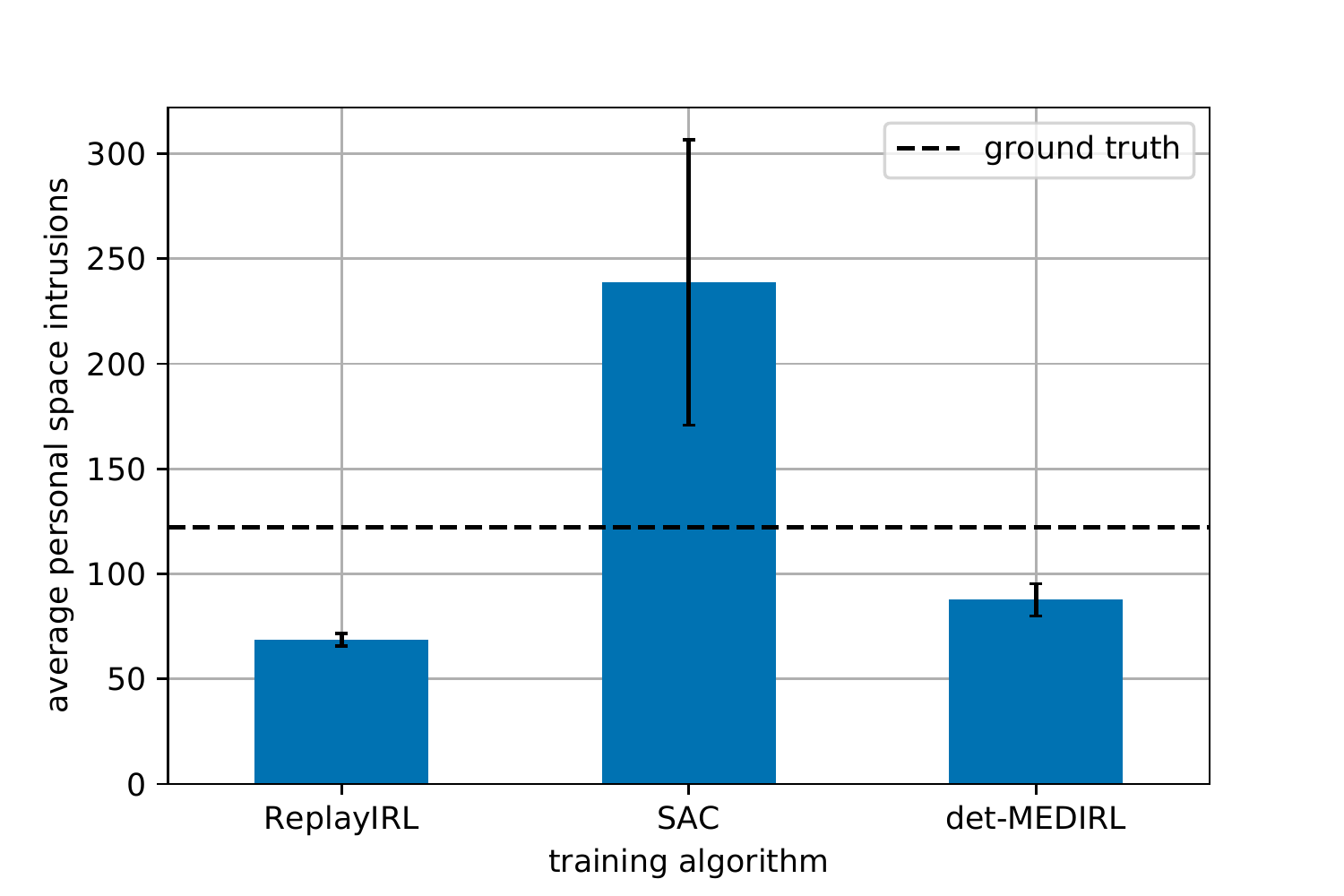}
    \caption{Zara subset.}
\end{subfigure}

\caption{Out of distribution intimate (a,b) and personal space (c,d) intrusions. The \ac{IRL} methods are both less intrusive than ground truth, indicating that a conservative policy for solitary agents was learned.}
\label{fig:proxemics}
\end{figure}

Results in \autoref{fig:proxemics} show that our method has the lowest proxemic intrusions out of all three compared methods on both test subsets, while \ac{SAC} significantly exceeds the expert ground truth. Interestingly, both \ac{det-MEDIRL} and ReplayIRL have lower proxemic intrusions than the ground truth, possibly because group behavior (which causes many such intrusions) is not considered by the solitary agents, which prefer a less intrusive policy to avoid the collision penalty.

\begin{figure}
\centering\centering
\begin{subfigure}[tbhp]{0.49\columnwidth}
    \includegraphics[width=\columnwidth]{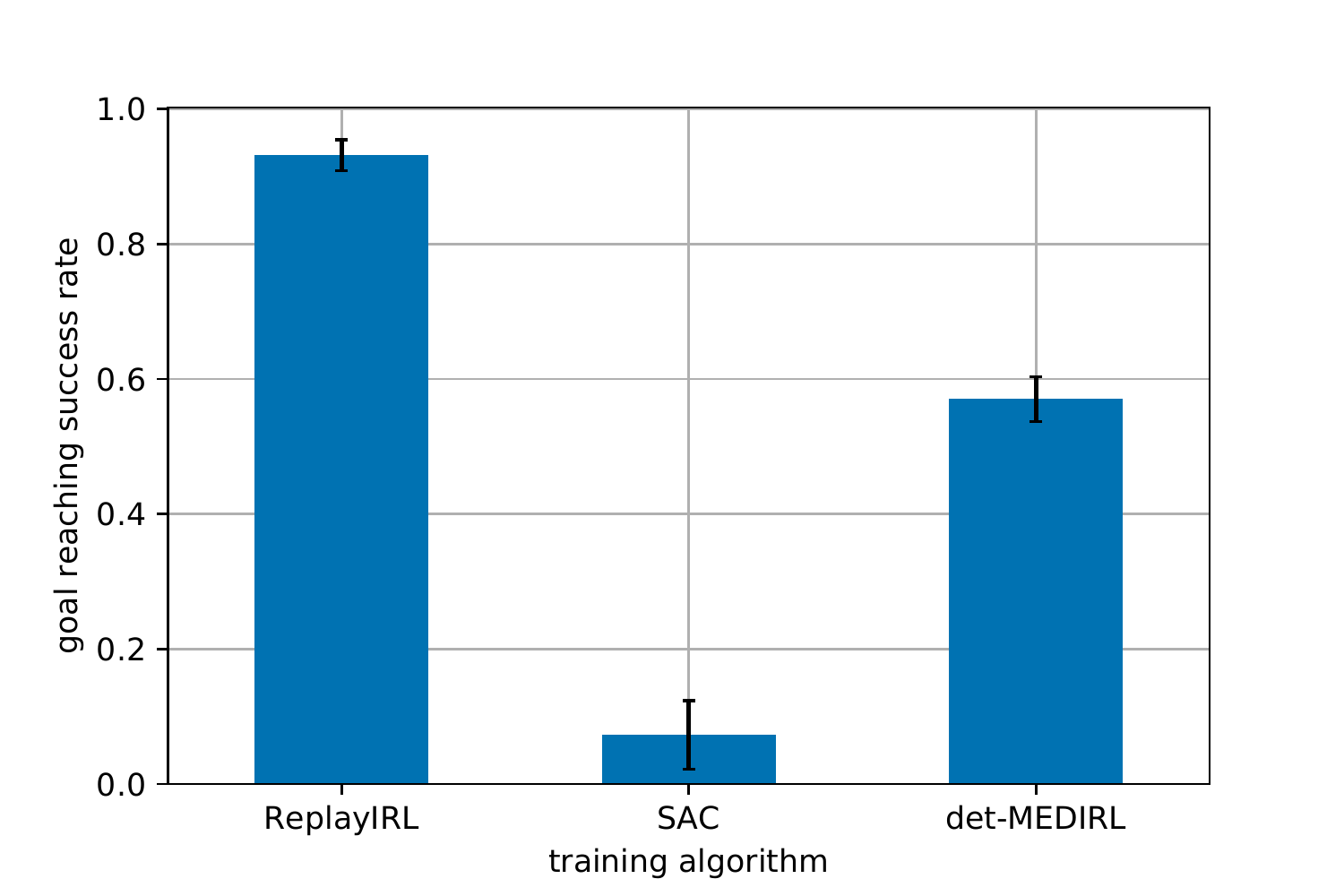}
    \caption{students003 subset.}
    \label{fig:student3-goal-reached}
\end{subfigure}
\hfill
\begin{subfigure}[tbhp]{0.49\columnwidth}
    \includegraphics[width=\columnwidth]{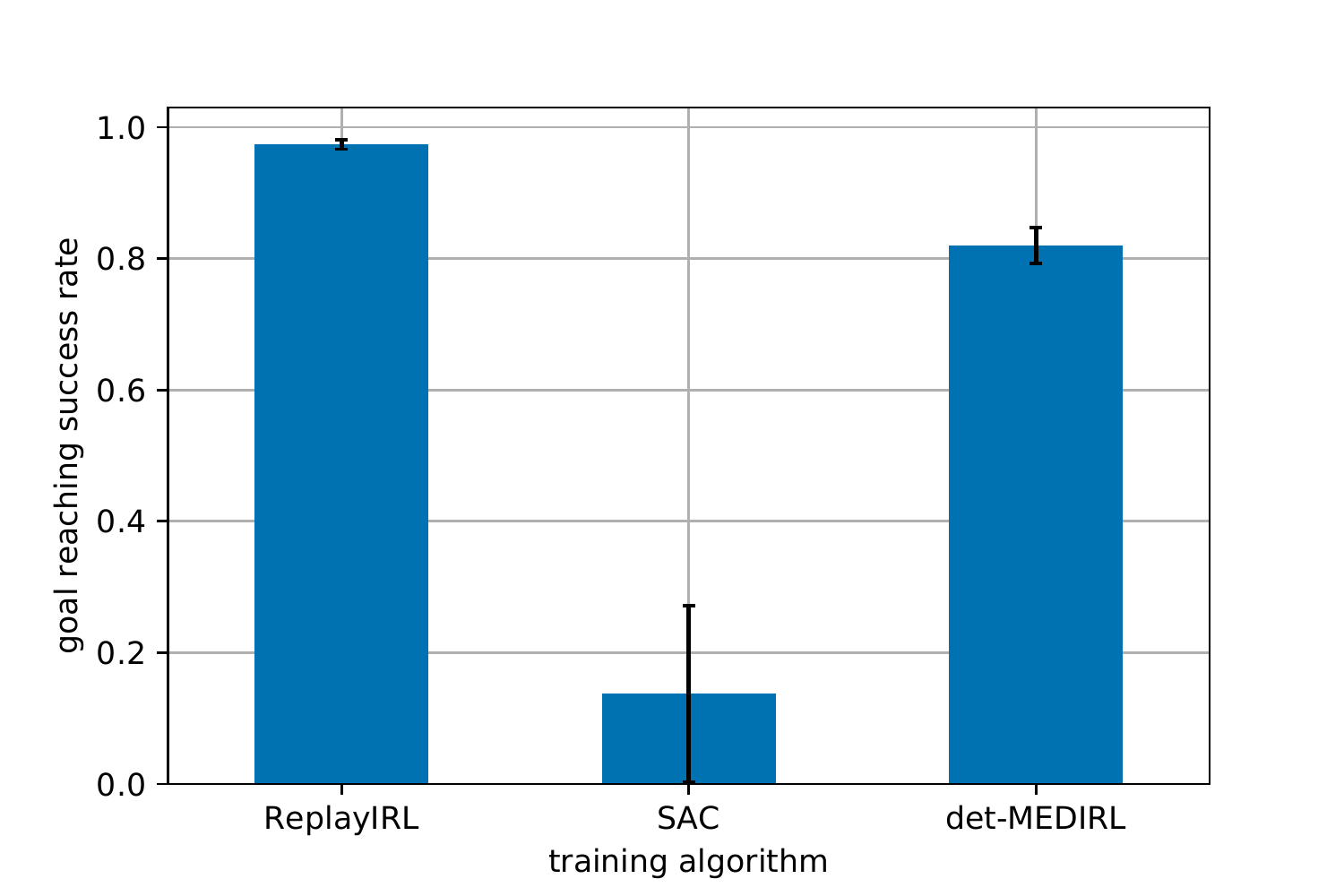}
    \caption{Zara subset.}
    \label{fig:zara-goal-reached}
\end{subfigure}

\caption{Out of distribution goal reaching success rate. ReplayIRL is better at collision-free navigation than \ac{det-MEDIRL} while \ac{SAC} fails to learn collision-free navigation.}
\label{fig:goal_reached}
\end{figure}

\begin{figure}
\centering\centering
\begin{subfigure}[tbhp]{0.49\columnwidth}
    \includegraphics[width=\columnwidth]{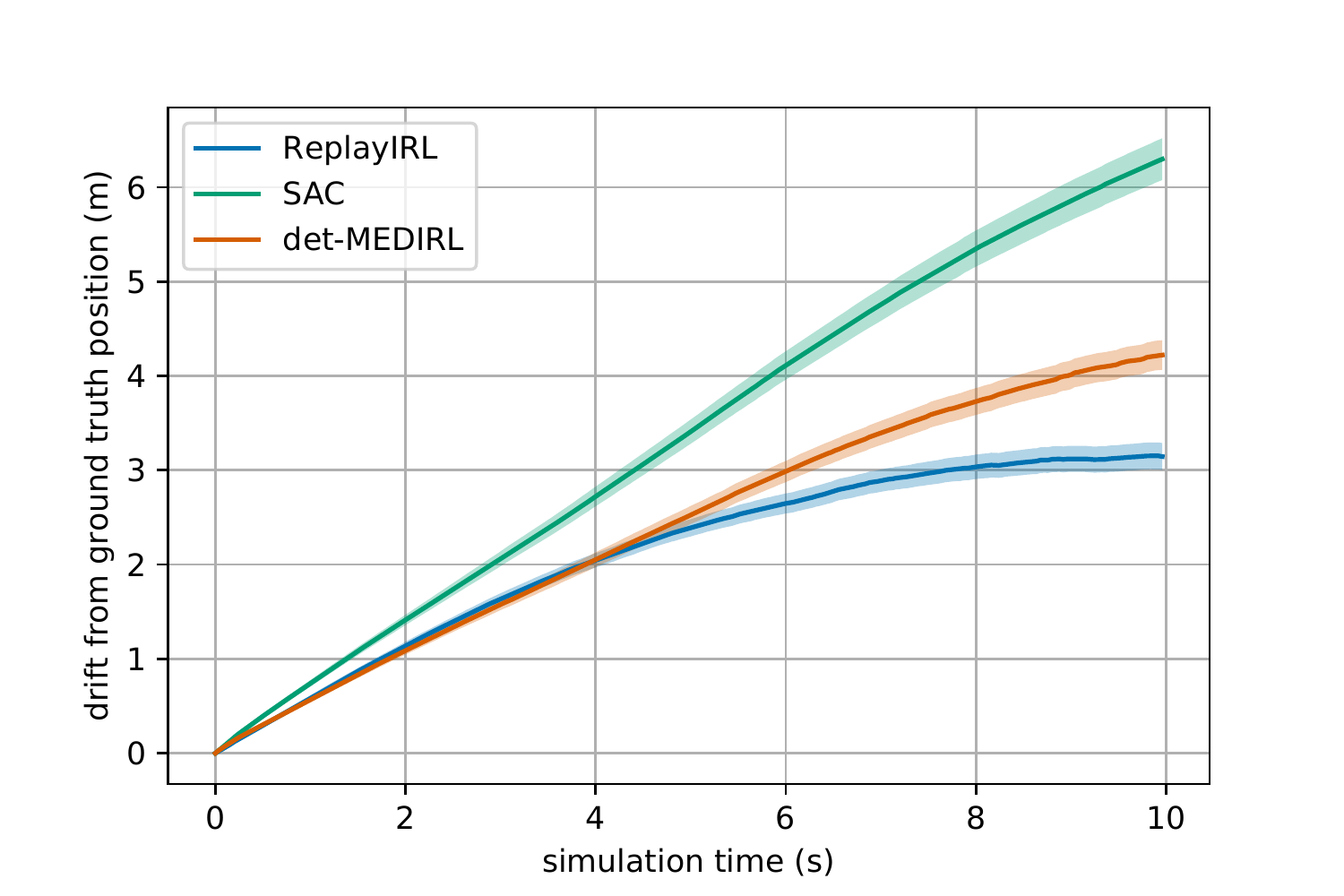}
    \caption{students003 subset.}
    \label{fig:irl-student-divergence}
\end{subfigure}
\hfill
\begin{subfigure}[tbhp]{0.49\columnwidth}
    \includegraphics[width=\columnwidth]{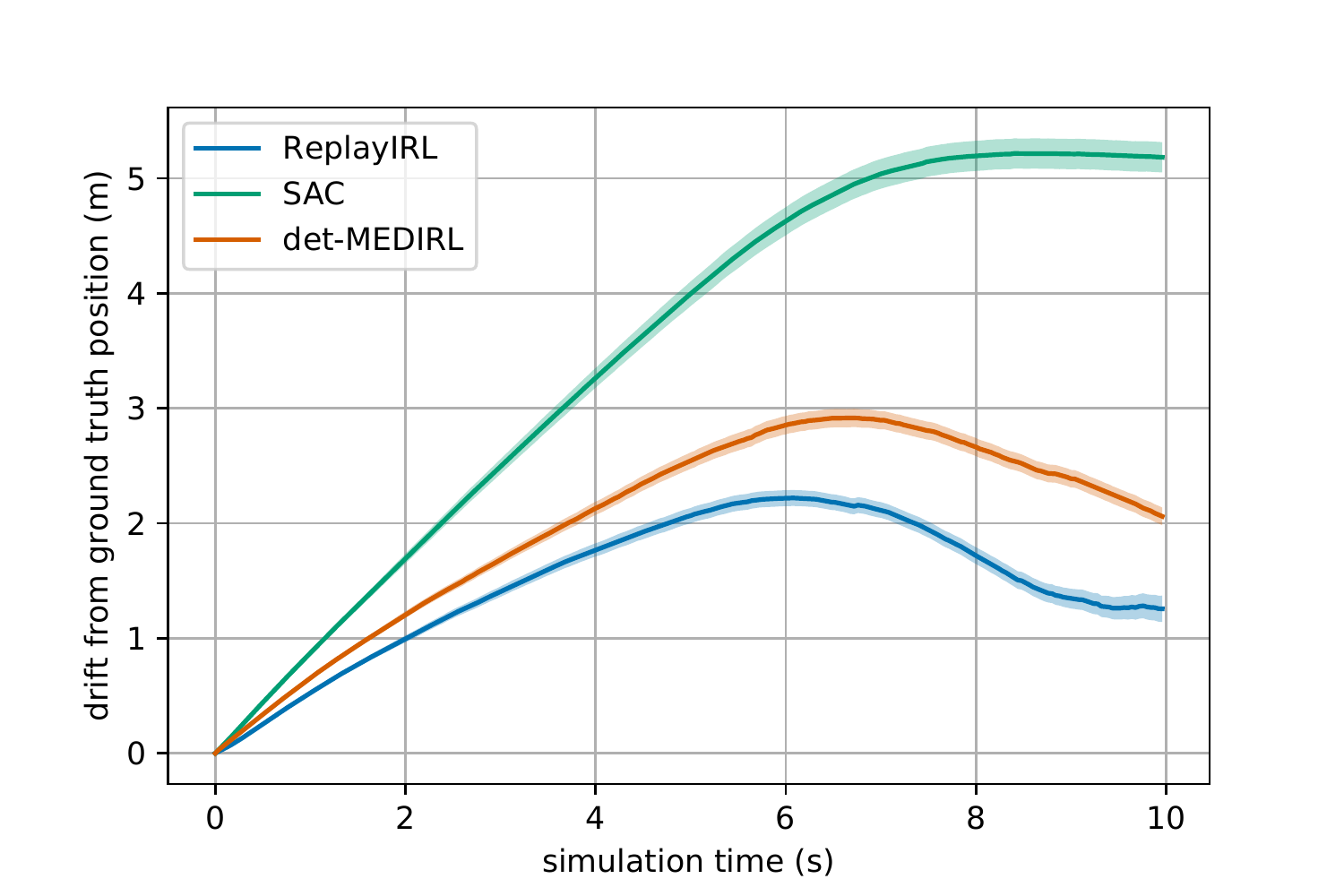}
    \caption{Zara subset.}
    \label{fig:irl-zara-divergence}
\end{subfigure}

\caption{Out of distribution average drift distance from ground truth position. ReplayIRL clearly stays closer to ground truth position than \ac{det-MEDIRL} while \ac{SAC} is completely divergent.}
\label{fig:divergence}
\end{figure}

\subsubsection{Drift from Ground Truth}

This measure calculates the average $l2$ distance from the ground truth position of the replaced pedestrian for the first $10\,s$ of simulation. Results in \autoref{fig:divergence} clearly illustrate that ReplayIRL stays closer to the ground truth position despite the limited nature of the risk features it uses. From \autoref{fig:irl-zara-divergence} at roughly $t=6\,s$ we can see that \ac{det-MEDIRL} and ReplayIRL are influenced similarly by the configuration of surrounding pedestrians indicating some similarity in the extracted reward in \ac{IRL} methods, while \ac{SAC} is completely divergent.

\subsubsection{Goal Reaching Success Rate} 
This metric measures the basic successful navigation and is defined as reaching the target goal location without collision with any other pedestrian. Results in \autoref{fig:goal_reached} illustrate that ReplayIRL is convincingly superior to \ac{det-MEDIRL} in terms of performance, while \ac{SAC} seems to fail to learn using hand-crafted rewards. 
This shows that the learned reward function is beneficially shaped to accelerate the \ac{RL} learning process.
We leave the investigation of this interesting \ac{RL} acceleration effect to future work.

While \ac{det-MEDIRL} achieves slightly better performance than the original implementation \cite{konarLearningGoalConditioned2021}, ReplayIRL performs significantly better in both the students003 and zara subsets, which can be attributed to an improved reward function due to importance sampling and accelerated training speed which leads to more \ac{IRL} learning iterations. 

\subsection{Sample Efficiency Results} \label{subsec:sample-efficiency-results}

\subsubsection{Feature Matching} \label{subsub:feature-matching}
\begin{figure}
\centering
\includegraphics[width=0.78\columnwidth]{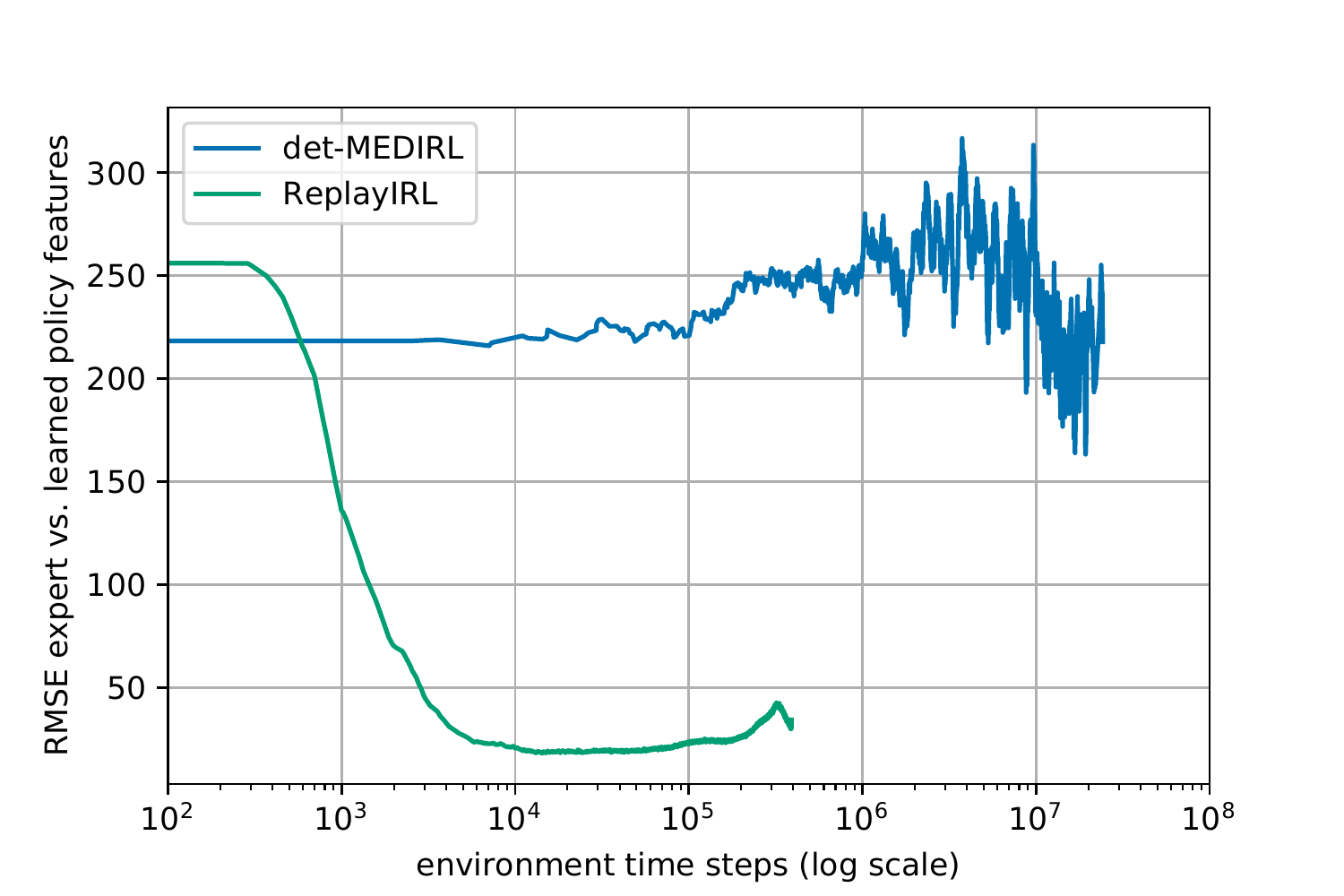}

\caption{\ac{RMSE} between average expert and replay buffer samples, 
measuring feature matching capability. The lower error shows better matching and the 
more rapid convergence of ReplayIRL is evident. 
\ac{det-MEDIRL} oscillations are indicative of a high variance objective function estimator.
}
\label{fig:irl-rmse-features}
\end{figure}

We record the \ac{RMSE} between the average expert state vector and the average policy state vectors at each training step as a measure of similarity between the agent and expert trajectories.
For ReplayIRL, the trajectory sample from the replay buffer is used for this comparison. For presentation clarity, we use a moving average with a window size of 100 when visualizing results in \autoref{fig:irl-rmse-features}.
The results in \autoref{fig:irl-rmse-features} illustrate
that ReplayIRL converges to an \ac{RMSE} minimum in $\sim10^5$ environment interactions, while \ac{det-MEDIRL} has just begun to converge at $10^7$ environment interactions. 
This achieves a two order of magnitude reduction in the number of required environment interactions. 
While we leave the application of ReplayIRL on a robotic platform to future work, these results reinforce the practicality of such an endeavor. 

We attribute this stark difference in sample efficiency to the density of \ac{IRL} updates in ReplayIRL; the training speed and sample efficiency are greatly increased as the \ac{IRL} subroutine does not require environment interaction. Note that while sample efficiency is more important for robot training, the training speed is also dramatically improved, down from 28.4 hours for \ac{det-MEDIRL} to 7.2 hours for ReplayIRL as measured by wall time.

\subsubsection{\ac{IRL} iterations vs. Environment Interactions}
\begin{figure}
\centering
\includegraphics[width=0.78\columnwidth]{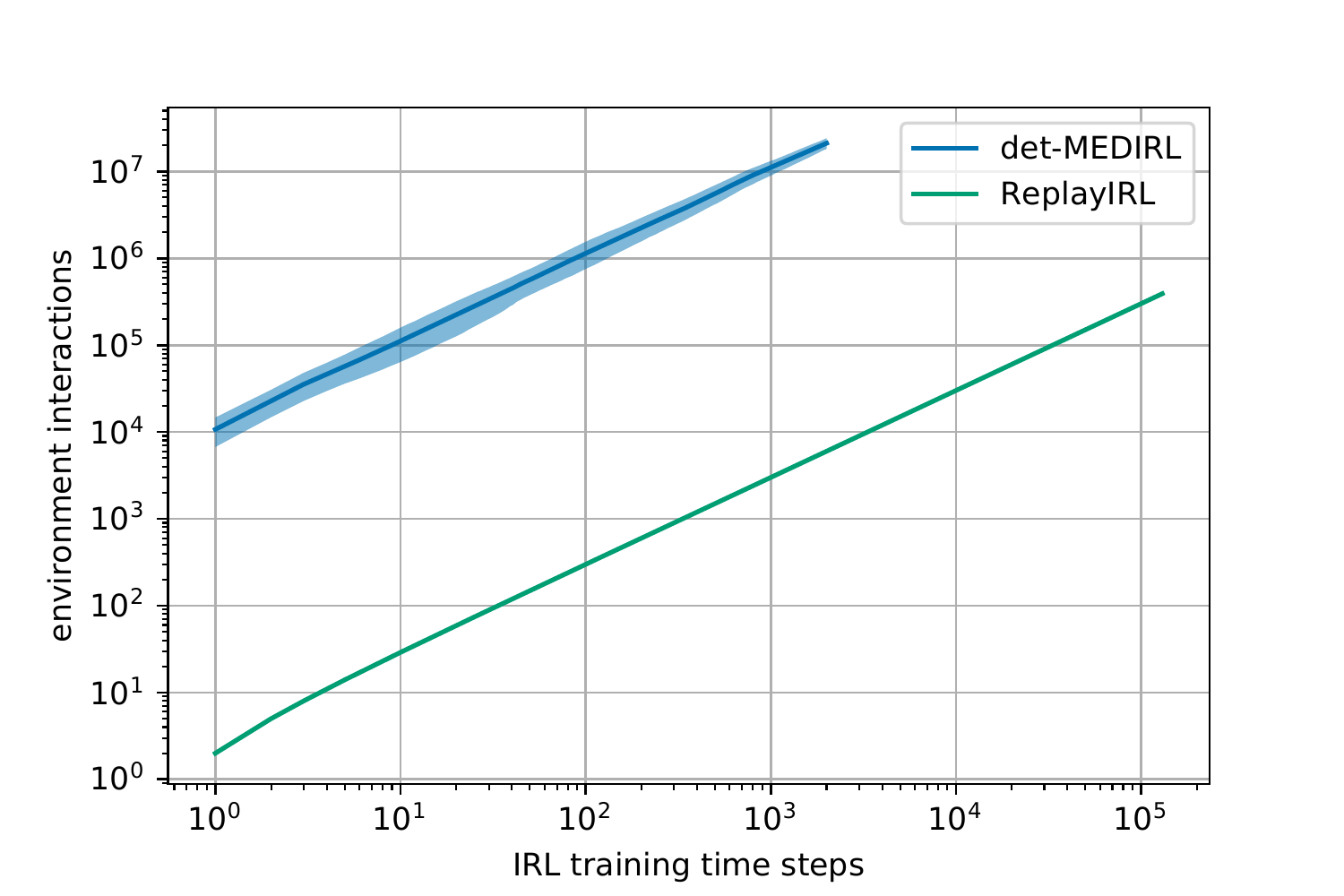}

\caption{ Environment interactions vs. training time step for \ac{IRL} methods. Note the log-log axis implies a much faster growth rate for det-MEDIRL.
Shaded region for \ac{det-MEDIRL} represents variance among all seeds.
}
\label{fig:irl-vs-env-steps}
\end{figure}

We record the number of environment interactions per \ac{IRL} iteration during training. A comparison of this metric provides us a direct measure of sample complexity as training progresses. The results, as seen in \autoref{fig:irl-vs-env-steps}, show the precipitous growth of the number of environment interactions required by \ac{det-MEDIRL} as opposed to ReplayIRL. Note that there is no uncertainty in the number of \ac{IRL} time steps vs. environment interactions as the training follows a set ratio. 

\section{CONCLUSIONS}
In this paper, we considered the important problem of robotic social navigation, which will play a crucial role in the integration of mobile robotics into previously human-inhabited social spaces. 
We presented ReplayIRL, an \ac{IRL} approach to social navigation which we demonstrate is more sample efficient, more performant, and faster to train than alternative \ac{IRL} based methods. 
In addition to this, future work might tackle methods of expert trajectory collection, feature extraction using commonly available robotic sensors, as well as improving the data efficiency of the algorithm.

\bibliographystyle{IEEEtran}
\bibliography{IEEEabrv,references.bib}

\end{document}